\documentclass[10pt,journal]{IEEEtranTCOM}

\usepackage[english]{babel}
\usepackage[usenames]{color}
\usepackage[cp1250]{inputenc}
\usepackage{amsfonts}
\usepackage{amsthm}
\usepackage{graphicx}
\usepackage{epsfig}
\usepackage{mathrsfs}
\usepackage{amsmath}
\usepackage{algorithm}
\usepackage{algorithmic}

\theoremstyle{plain}
\newtheorem{df}{Definition}

\begin{document}
\newcommand{\bea}{\begin{eqnarray}}
\newcommand{\eea}{\end{eqnarray}}
\newcommand{\be}{\begin{equation}}
\newcommand{\ee}{\end{equation}}
\newcommand{\beas}{\begin{eqnarray*}}
\newcommand{\eeas}{\end{eqnarray*}}
\newcommand{\bs}{\backslash}
\newcommand{\bc}{\begin{center}}
\newcommand{\ec}{\end{center}}
\def\SC {\mathscr{C}}

\title{Rapid parametric density estimation}
\author{\IEEEauthorblockN{Jarek Duda}\\
\IEEEauthorblockA{Jagiellonian University,
Golebia 24, 31-007 Krakow, Poland,
Email: \emph{dudajar@gmail.com}}}
\maketitle

\begin{abstract}
Parametric density estimation, for example as Gaussian distribution, is the base of the field of statistics. Machine learning requires inexpensive estimation of much more complex densities, and the basic approach is relatively costly maximum likelihood estimation (MLE). There will be discussed inexpensive density estimation, for example literally fitting a polynomial (or Fourier series) to the sample, which coefficients are calculated by just averaging monomials (or sine/cosine) over the sample. Another discussed basic application is fitting distortion to some standard distribution like Gaussian - analogously to ICA, but additionally allowing to reconstruct the disturbed density. Finally, by using weighted average, it can be also applied for estimation of non-probabilistic densities, like modelling mass distribution, or for various clustering problems by using negative (or complex) weights: fitting a function which sign (or argument) determines clusters. The estimated parameters are approaching the optimal values with error dropping like $1/\sqrt{n}$, where $n$ is the sample size.
\end{abstract}
\section{Introduction}
\begin{figure}[t!]
    \centering
        \includegraphics{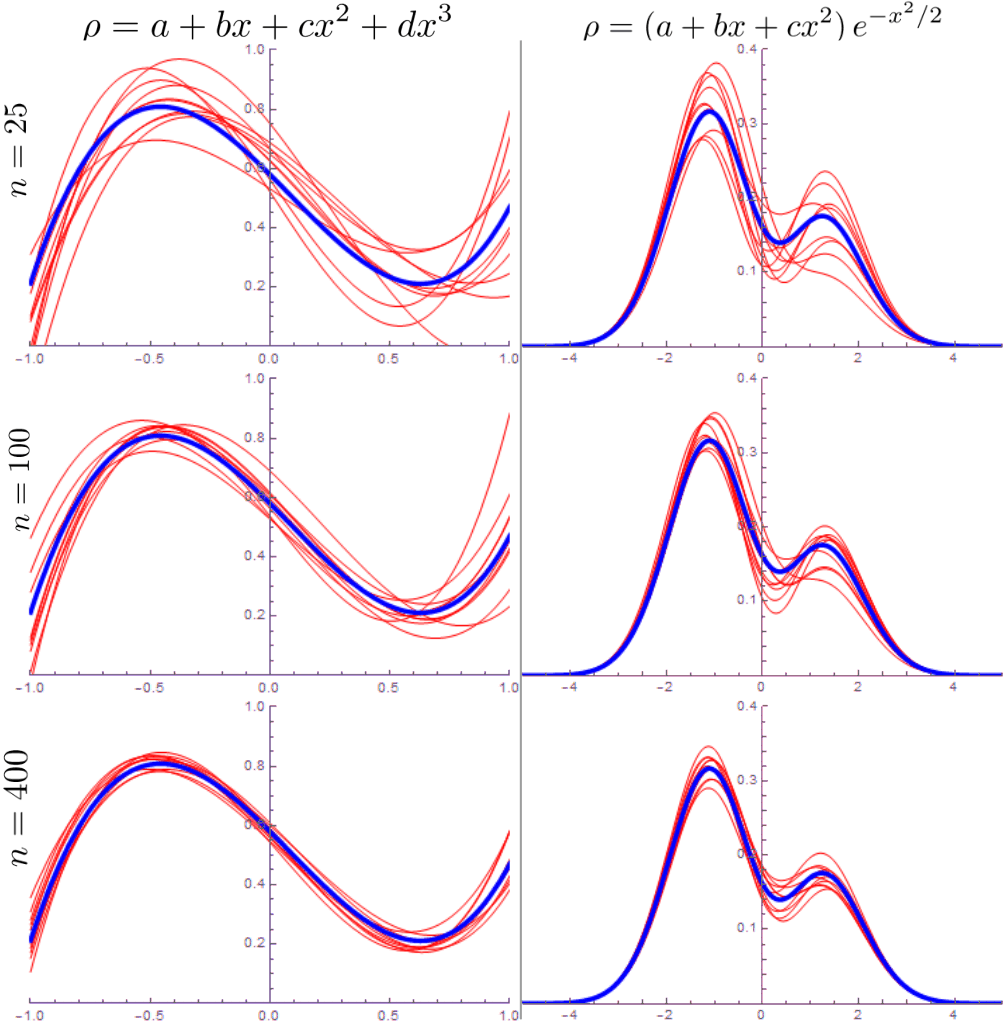}
        \caption{Two basic discussed examples of fitting parameters for assumed family of probability densities: polynomials (left column) in $[-1,1]$ range and polynomials multiplied by $e^{-x^2/2}$ (right) in $\mathbb{R}$, basing on a random sample of size $n=25$ (top row), $n=100$ (middle) or $n=400$ values (bottom), generated using the assumed probability distribution - represented as the thick blue line. Every plot contains also 10 thin red lines representing results of 10 independent experiments of estimating the parameters basing on the obtained size $n$ sample. Inaccuracy drops with $1/\sqrt{n}$, what can be seen in dispersion dropping approximately twice every row. For convenience there will be discussed using orthogonal family of functions, for example polynomials, making their parameters independent - calculated as just average of the value of a given function over the obtained sample. The polynomial formula used for the left column can be also expressed as $\rho=\frac{5}{8}\left(1-3x^2+[x](15x-21x^3)+[x^2](9x^2-3)+[x^3](35x^3-21x)\right)$,
        where $[f]$ denotes average of function $f$ over the sample.}
       \label{overv}
\end{figure}
\noindent The complexity of our world makes it useful to imagine a sample obtained from some measurements, in nearly all kind of science, as coming from some probability distribution. The natural approach to model this distribution is using a parametric function, which parameters should be estimated basing on the sample. The best known example is Gaussian (normal) distribution, which parameters are estimated from averages over the sample of all up to degree 2 monomials.

To model real data we usually need more complex densities. In machine learning there are popular kernel density estimators (KDE)~(\cite{ker1,ker2}), which smoothen the sample by convolving it with a kernel: a nonnegative function integrating to 1, for instance a Gaussian distribution. However, one issue is that it requires arbitrarily choosing the width of this kernel: if it is too narrow we get a series of spikes, if too wide we loose the details. Another problem is that such estimated density is a sum of potentially large number of functions - is very costly to directly work with.

Hence, usually it would be more convenient to have a parametric model with a relatively small number of parameters, for example to locally approximate density with a polynomial. The standard approach to estimate such parameters is the maximum likelihood estimation~(MLE)~(\cite{MLE1,MLE2}), which finds the parameters maximizing likelihood of obtaining the given sample. However, beside really simple examples, such maximization would require costly numerical procedures like gradient descent.\\

\begin{figure}[t!]
    \centering
        \includegraphics{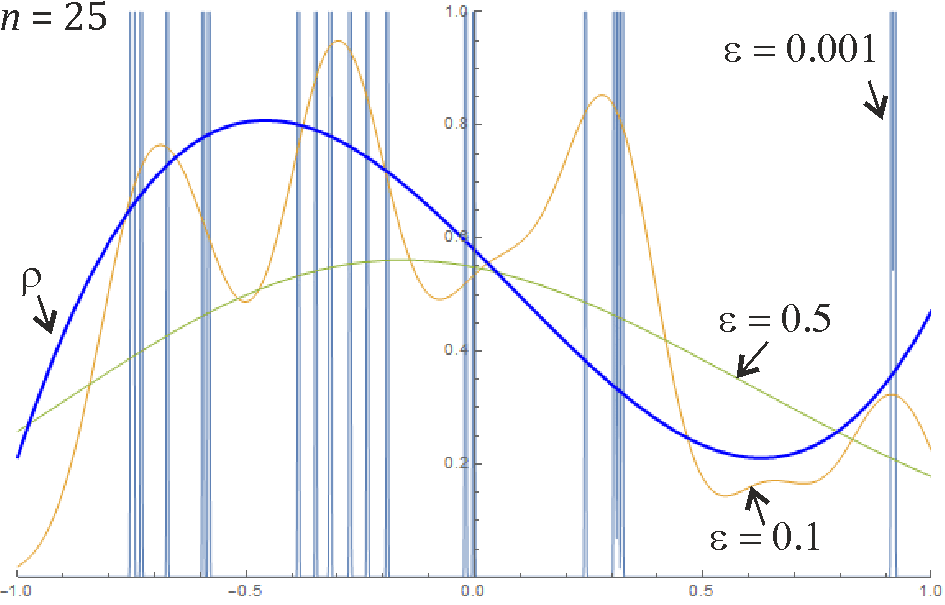}
        \caption{Examples of smoothing (KDE) sample of $n=25$ points from density represented by the thick blue line. The three thin lines  represent smoothing with Gaussian kernel for standard deviation $\epsilon=0.5,\ 0.1$ and $0.001$. We can fit for example a polynomial to such smoothen sample. Large $\epsilon$ intuitively blurs the information, so we focus here on mean-square fitting to the sharpest: $\epsilon\to 0$ limit spikes, sum of Dirac deltas.}
       \label{spike}
\end{figure}

As in examples from fig. \ref{overv}, we will discuss here inexpensive parametric estimation, especially with a linear combination of a chosen family of functions, for instance with a polynomial or Fourier series in a finite region. It is based on mean-square fitting of the optimal parameters to the sample smoothed with a kernel (KDE). Surprisingly, the mean-square ($L^2$) fitting gives the best agreement for the 0 width limit of the kernel: degenerated to Dirac delta, like in fig. \ref{spike}, by the way removing the very inconvenient requirement of arbitrarily choosing the kernel width. Intuitively, it fits a function to a series of spikes, what makes it much simpler to calculate and asymptotically leads to the real parameters of the used density: the error drops like $1/\sqrt{n}$ with $n$ being sample size. 

The parameters of a linear combination are obtained analogously to algebraic projection on a preferably (complete) orthogonal base of functions, making the estimated coefficient for a given function as just the average of value of this function over the sample. For fitting order $m$ polynomial in a finite region (preferably a hyperrectangle: product of ranges), we just need to calculate averages over the sample of all monomials of power up to $m$ (estimators of moments). Orthogonality of the used family allows to ask independently about each coefficient, what can be used to directly work with as high order polynomial approximation as needed, eventually neglecting coefficients which turn out close to zero. Finally, the Stone-Weierstrass theorem theorem says that any continuous function on a closed interval can be uniformly approximated as closely as desired by polynomials, making it theoretically possible to asymptotically recreate any continuous density using the discussed approach. Analogously for Fourier series, where the coefficients are obtained as just averages of corresponding sine or cosine over the sample.

Beside being much less expensive to calculate than MLE, minimization of mean-square error might lead to more appropriate densities for some machine learning problems. Specifically, MLE categorically requires all probabilities being positive on the sample, what seems a strong restriction for example for fitting polynomials. Mean-square optimization sometimes leads to negative values of density, what obviously may bring some issues, but can also lead to better global agreement of the fitted function.

Another discussed basic application, beside literally fitting polynomial to a sample in a finite region, is estimating and describing distortion from some simple expected approximated distribution, like uniform or Gaussian distribution. For example for testing long range correlations of a pseudorandom number generators (PRNG). Taking $D$ length sequences of its values as points from $[0,1]^D$ hypercube, ideally they should come from $\rho=1$ uniform density. The discussed method allows to choose some testing function (preferably orthogonal to $\rho=1$: integrating to 0) describing our suspicion of distortion from this uniform density, like $\prod_i (x_i-1/2)$, and estimate its coefficient basing on the sample - test if the average of its values approaches zero as expected.

More important example is modelling distortions form the Gaussian distribution. The standard way is to calculate higher (central) moments, however, it is a difficult problem to translate them back into the actual probability density - so called "the problem of moments"~\cite{mom}.

The most widely used methods for understanding distortion from Gaussian distribution, as noise in which we would often like to find the real signal, is probably the large family of independent component analysis (ICA)~\cite{ica} methods. They usually start with normalization: shifting to the mean value, and rescaling directions of eigenvectors of the covariance matrix accordingly to the corresponding eigenvalues, getting normalized sample: with unitary covariance matrix. Then it searches for distortions from the perfect Gaussian (as noise) of this normalized sample, for example as directions maximizing kurtosis - candidates for the real signal. The discussed method allows to directly fit parameters of distortion as a linear combination of some chosen family of functions, like polynomials multiplied by $e^{-x^2/2}$. Its advantages comparing to ICA is the possibility to model multiple modes for every direction (like degrees of polynomial) and allowing to reconstruct the disturbed density function. It can also work with different than Gaussian types of tail, like $e^{-|x|}$ or $1/|x|^K$. \\

While the main focus here are probabilistic densities: nonnegative and integrating to 1, this approach can be also used to model more general or abstract densities: fitting a relatively simple function like a polynomial to a smoothen set of points. For example, the original motivation for this approach was fitting a low order polynomial to a distribution of atomic masses or electron negativity along the longest axis of a chemical molecule~\cite{mole}. Coefficients of this polynomial describe some prosperities of this molecule and can be used as a part of information in its fingerprint (descriptor) for virtual screening. Another basic application can be trend-seeking estimation of probability of a symbol basing on its previous appearances in adaptive data compressors: exploiting trends like rising or dropping probability by fitting and extrapolating a polynomial.

Finally, this approach can be also applied for various clustering problems. By thresholding the estimated density function we can determine regions of relatively high density, interpreted as clusters for unsupervised learning. For supervised problem: with some additional information about classification of points into clusters, we can use these labels as weights while averaging. For two classes we can use weights of opposite signs to distinguish them (under-represented class should have correspondingly higher absolute weights), then the sign of the fitted density can be used to determine the cluster. We can also directly distinguish a larger number of classes with a single function, for example by using complex (or vector) weights (like $W=\pm 1 \pm \sqrt{-1}$), then use its argument to determine the cluster (like $\lfloor \frac{2}{\pi}\arg(\rho)\rfloor$).

The standard approaches in machine learning, like neural networks or SVMs, are based on linear classifiers with parameters found e.g. by backpropagation. The discussed approach not only allows to generalize it to directly use for example higher order polynomials here, but also directly fits their parameters by just averaging over the sample. Generalization to higher order polynomials with clusters defined for example as thresholded polynomials (semi-algebraic sets), gives much stronger separability strength. For example sign of $xy$ monomial already classifies the XOR pattern. Paroboloida as second order polynomial can directly separate an ellipsoid, and much more for higher order polynomials. This approach can be used for example as one of layers of neural network with initially directly fitted coefficients, then improved e.g. through backpropagation. Due to simplicity of calculating the coefficients, the formulas for fitting them can be also directly put into the backpropagation process.
\section{General case}
\noindent Imagine we have a sample of $n$ points in $D$ dimensional space: $S=\{\textbf{x}^i\}_{i=1..n}$, where $\textbf{x}^i =\{x^i_1,\ldots,x^i_D\}\in \mathbb{R}^D$. Assuming it comes from some probability distribution, we will try to approximate (estimate) its probability density function (PDF) as a function from some chosen parameterized family $\rho_\textbf{a}:\mathbb{R}^D\to \mathbb{R}$ for some $m$ parameters $\textbf{a}=\{a_1,\ldots,a_m\}$. In other words, we would like to estimate the real parameters basing on the observed sample. For generality, assume each point has also a weight $W(\textbf{x})$, which is $W=1$ for probability estimation, but generally it can represent a mass of an object, or can be negative for separation into two classes, or even be a complex number or a vector for example for simultaneous classification into multiple classes.

Preferably, a probabilistic $\rho_\textbf{a}$ should be smooth, nonengative and integrate to one. However, as we will mostly focus on $\rho_\textbf{a}$ being a linear combination, in some cases there will appear negative values as an artifact of the method. Probabilistic normalization (integration to 1) will be enforced by construction or additional constraints.

A natural approach is smoothing the sample (before fitting) into a continuous function, for example by convolution with a kernel (KDE):
\be g_\epsilon(\textbf{x}) := \frac{1}{n}\sum_{\textbf{y}\in S} W(\textbf{y})\, k_{\epsilon}(\textbf{x}-\textbf{y})\ee

\noindent where $k$ is nonnegative and integrates to 1, for example is the Gaussian kernel: $k_\epsilon(\textbf{x}) = (2\pi \epsilon^2)^{-D/2}\, e^{-\textbf{x}\cdot \textbf{x}/2\epsilon^2}$. Obviously, for probabilistic $W=1$, $g_\epsilon$ is also nonnegative and integrates to 1.\\

We can now formulate the basic problem as finding parameters minimizing some distance from $g_\epsilon$:
\be \min_{\textbf{a}} \| \rho_\textbf{a} - g_\epsilon \| \label{min} \ee

However, there remains a difficult question of choosing the parameter $\epsilon$. In fact it is even worse as the kernel assumed here is spherically symmetric, while for the real data more appropriate might be for example some elliptic kernel with a larger number of parameters to choose, which additionally should be able to vary with position.

Surprisingly, we can remove this dependency on $\epsilon$ by choosing mean-square norm ($L^2$), which  allows to perform the $\epsilon\to 0$ limit of kernel: to Dirac delta as in fig. \ref{spike}. The $g_\epsilon$ becomes a series of spikes in this limit, no longer being a type of function expected while fitting with a family of smooth functions. However, it turns out to well behave from the mathematical point of view. Intuitively, while $\epsilon\to \infty$ limit would mean smoothing the sample into a flat function - loosing all the details, the $\epsilon\to 0$ limit allows to exploit the sharpest possible picture.

Assume we would like to minimize mean-square norm $\|f\|:=\sqrt{\langle f,f \rangle }$ for scalar product:
\be \langle f,g \rangle:=\int_{\mathbb{R}^D} f(\textbf{x})\, g(\textbf{x})\,w(\textbf{x}) \,d\textbf{x}  \ee

\noindent where we will usually use constant weight $w=1$. However, for fitting only local behavior, it might be also worth to consider some vanishing $w\to 0$ while going away from the point of interest.

For mean-square norm, the minimization (\ref{min}) is equivalent to:
$$ \min_{\textbf{a}} \| \rho_\textbf{a} - g_\epsilon \|^2 = \min_{\textbf{a}} \langle \rho_\textbf{a} - g_\epsilon , \rho_\textbf{a} - g_\epsilon\rangle \ =$$
\be \min_{\textbf{a}}\ \|\rho_\textbf{a}\|^2 - 2 \langle g_\epsilon,\rho_\textbf{a} \rangle + \|g_\epsilon\|^2 \label{min1}\ee

In the $\epsilon\to 0$ limit we would have $\|g_\epsilon\|\to \infty$. However, as we are only interested in the optimal parameters $\textbf{a}$ here, and $\|g_\epsilon\|$ does not depend on them, we can just remove this term resolving the issue with infinity.  The term $ \langle g_\epsilon,\rho_\textbf{a} \rangle$ becomes $\frac{1}{n}\sum_{\textbf{x}\in S} w(\textbf{x})W(\textbf{x})\, \rho_\textbf{a}(\textbf{x})$ in the $\epsilon\to 0$ limit, finally getting:

\begin{df}
  The general considered minimization problem is
\end{df}
\be \min_{\textbf{a}}\ \langle \rho_\textbf{a}, \rho_\textbf{a}\rangle - \frac{2}{n} \sum_{\textbf{x}\in S} w(\textbf{x})W(\textbf{x})\,  \rho_\textbf{a}(\textbf{x}). \label{min2}\ee

Its necessary differential condition is:
\be  \langle \rho_\textbf{a}, \, \partial_{a_j} \rho_\textbf{a}\rangle = \frac{1}{n} \sum_{\textbf{x}\in S} w(\textbf{x})W(\textbf{x})\, (\partial_{a_j} \rho_\textbf{a})(\textbf{x})  \label{nec} \ee
for all $j\in\{1,\ldots,m\}$.

\section{Density estimation with \\a linear combination}
\noindent The most convenient application of the discussed method is density estimation with a linear combination of some family of functions, for instance polynomials or sines and cosines. In this and the following section we assume:

\be \rho_\textbf{a} = \sum_{i=1}^m a_i f_i \ee
for some functions $f_i: \mathbb{R}^D\to \mathbb{R}$ (not necessarily non-negative in the entire domain). 

While in practice we can rather only work on a finite family, it can be in fact an infinite complete orthogonal base - allowing to approximate any continuous function as close as needed, what is true for example for polynomials and Fourier base in a finite closed interval.

\subsection{Basic formula}
In the linear combination case, $\partial_{a_j} \rho_\textbf{a} = f_j$, and the necessary condition (\ref{nec}) becomes just:
\be \sum_i  a_i\, \langle f_i,f_j \rangle  = \frac{1}{n} \sum_{\textbf{x}\in S} w(\textbf{x})W(\textbf{x})\, f_j(\textbf{x})\ee
for $j=1,\ldots,m$. Denoting $(\langle f_i,f_j \rangle)$ as the $m\times m$ matrix of scalar products, the optimal coefficients become:

\be \textbf{a}^T = \, (\langle f_i,f_j \rangle)^{-1} \cdot  ([f_1],\ldots,[f_m])^T \label{gene}\ee
\be \textrm{where}\quad [f]:=\frac{1}{n} \sum_{\textbf{x}\in S} w(\textbf{x})W(\textbf{x})\, f(\textbf{x}) \ee
is estimator of the expected value of $f$. Assuming orthonormal set of functions: $\langle f_i,f_j \rangle  = \delta_{ij}$ and $w=W=1$ standard weights, we get a surprisingly simple formula:
\be a_i = [f_i] = \frac{1}{n} \sum_{\textbf{x}\in S} f_i(\textbf{x}) \label{ai}\ee

\be \rho_{\textbf{a}} (\textbf{x}) = \sum_i\, [f_i]\, f_i(\textbf{x}) \label{rho} \ee
We see that using an orthonormal family of functions is very convenient as it allows to calculate estimated coefficients independently, each one being just (weighted) average over the sample of the corresponding function. It is analogous to making algebraic projections of the sample on some orthonormal base. The independence allows this base to be potentially infinite, preferably complete to allow for approximation of any function, for example a base of orthogonal polynomials. Orthonormality (independence) allows to inexpensively estimate with as high order polynomial as needed. Eventually, without orthonormality, there can be used the more general formula (\ref{gene}) instead.

As the estimated coefficients are just (weighted) average of functions over the sample, it can be naturally generalized to continuous samples, for example representing some arbitrary knowledge, where the average can be calculated by integration.

\be a_i = [f_i]=\frac{1}{C} \int_S w(\textbf{x})W(\textbf{x})\,f(\textbf{x}) d\textbf{x}  \ee
with some normalization if needed, for example $C=\int_S w(\textbf{x})W(\textbf{x}) d\textbf{x}$. For clustering applications there can be used $C=1$. We can also mix some continuous arbitrary knowledge with sample of points from measurements by treating them as Dirac deltas to combine sum with integrations.

\begin{figure}[t!]
    \centering
        \includegraphics{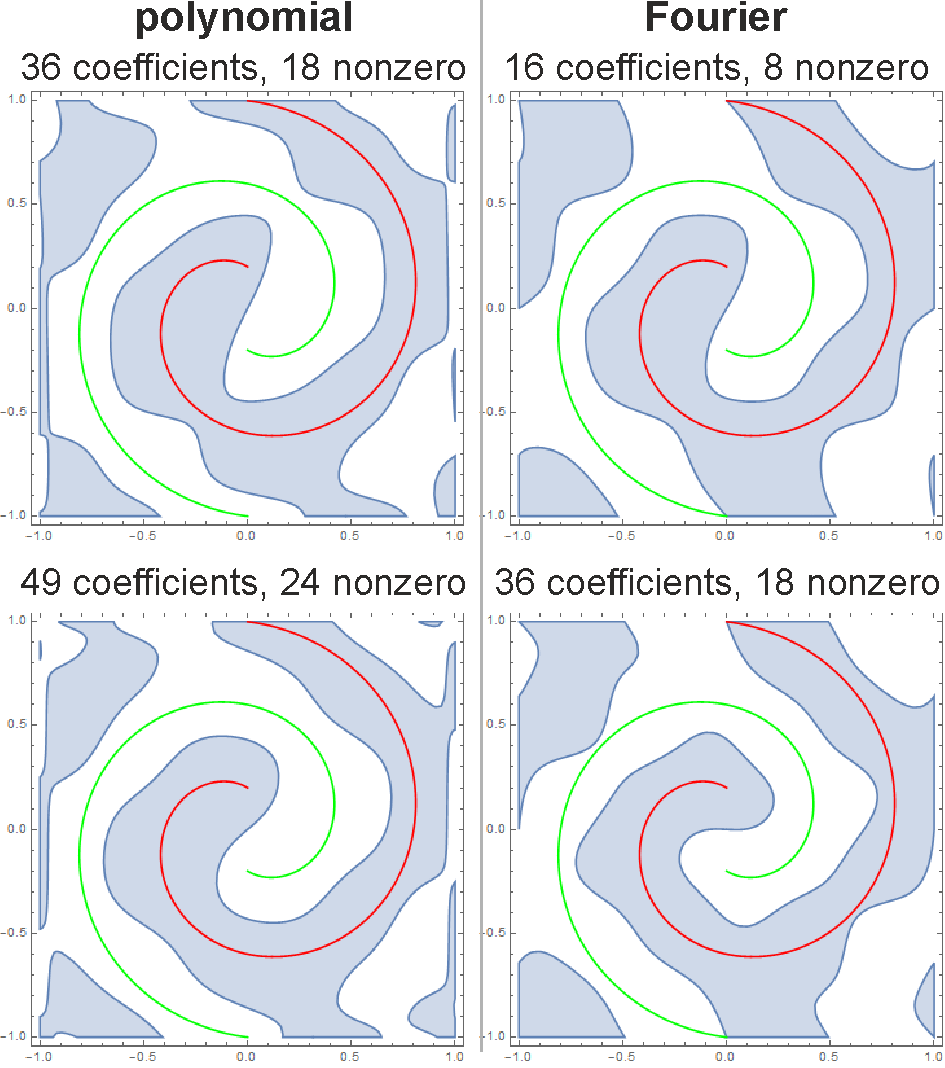}
        \caption{Examples of generalization to separate two regions (clusters) in 2D: defined by the red and green spiral. One of them was used with weight $W=1$, the second with $W=-1$ to fit order 6 (top) or 7 (bottom) polynomial (left), or Fourier base (right): $(\sin(im),\cos(im))$ for $i$ up to 2 (top) or 3 (bottom). The marked blue region was determined by $\rho>0$ condition. As there was used orthonormal family of functions, each coefficient is just average of a given function. This time instead of discrete samples, we have some arbitrary knowledge given by continuous sets (spirals): the average was made by integration. Due to symmetries, most of coefficients have turned out zero here as specified in the figure, generally suggesting to neglect functions with low coefficient in the considered linear combination. While it properly reconstructs complex boundaries, we can also see introduced some artifacts far from the samples.}
       \label{2spir}
\end{figure}

\begin{figure}[t!]
    \centering
        \includegraphics{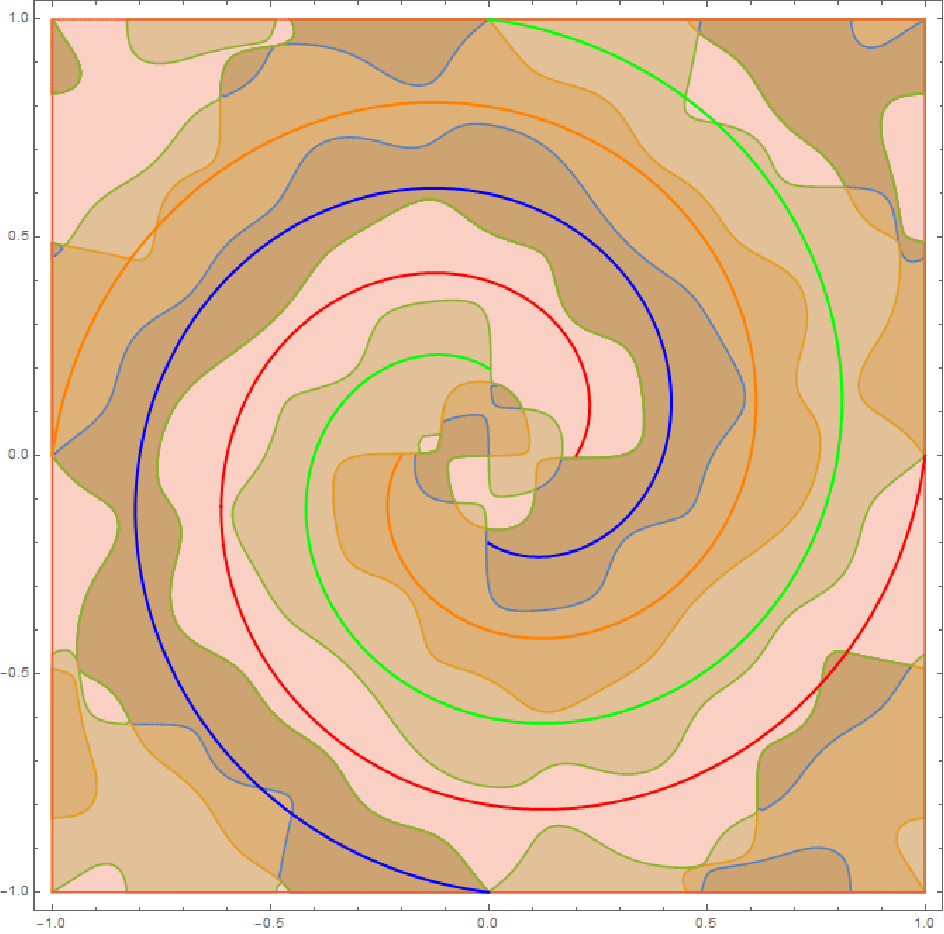}
        \caption{Example of generalization of approach from fig. \ref{2spir} to simultaneously separate a larger number of classes - this time into 4 classes defined by arbitrary knowledge as the 4 spirals of different colors. Instead of using $\pm 1$ weights as previously, this time there were used 4 complex weights: $W=\pm 1 \pm \sqrt{-1}$. Then classification was made based on the argument (angle) of the fitted linear combination: $\lfloor \frac{2}{\pi}\arg(\rho)\rfloor$ (in which of 4 quaters this function is). There was used Fourier base for $(\sin{im},\cos(im))$ for $i$ up to 5, getting 200 real coefficients, 100 of them are nonzero.}
       \label{4class}
\end{figure}

Figures \ref{2spir} and \ref{4class} shows such examples for using spirals as continuous arbitrary knowledge. They also use  $W=\pm 1$ or $W=\pm 1 \pm \sqrt{-1}$ sample weights to determine the clusters basing on the sign of the fitted function, or its argument in the complex case.

\subsection{Asymptotic behavior ($n\to \infty$)}
Assume $W=w=1$ weights, orthonormal family of functions $(\langle f_i,f_j\rangle = \delta_{ij})$ and that the real density can be indeed written as $\rho=\sum_i a_i f_i$ (not necessarily finite).

For estimation of $a_i$, the (\ref{rho}) formula says to use $a_i\approx [f_i]$: average value of $f_i$ over the obtained sample. With the number of points going to infinity, it approaches the expected value of $f_i$ for this probability distribution:

$$ [f_i] \xrightarrow{n\to\infty} \int f_i\, \rho\, d\textbf{x} =\left\langle f_i,\, \sum_i a_i f_i\right\rangle=a_i $$

\noindent As needed, $[f_i]$ approaches the exact value of $a_i$ from the assumed probability distribution. From the Central Limit Theorem, the error of $i$-th coefficient comes from approximately a normal distribution of width being standard deviation of $f_i$ (assuming it is finite) divided by $\sqrt{n}$:
\be [f_i]-a_i \sim \mathcal{N}\left(0,\frac{1}{\sqrt{n}}\sqrt{\int (f_i-a_i)^2 \rho\, d\textbf{x}}\right) \ee
Hence, to obtain twice smaller errors, we need to sample 4 times more points.

\subsection{Nondegenerate kernel corrections}
The zero width kernel limit (to Dirac delta) were only used to replace scalar product with sum while going from (\ref{min1}) to (\ref{min2}). For a general kernel $k$ (nonegative, integrating to 1), orthonormal base and $D=1$ case, we would analogously obtain estimation:

$$ a_i =\frac{1}{n}\int \sum_{x\in S} k(y-x)\, f_i(y) dy\approx $$
$$\approx\frac{1}{n}\sum_{x\in S} \left(f_i(x)+\frac{1}{2}f''_i(x)\int h^2 k(h)dh \right) $$

\noindent where we have used second order Taylor expansion
$\left(f_i(x+h)\approx f_i(x)+h\,f'_i(x)+\frac{1}{2}h^2\,f''_i(x)\right)$ and assumed symmetry of kernel $(k(-h)=k(h))$, zeroing the first order correction. Finally we got the second order correction:
\be a_i \approx [f_i + v f''_i]\qquad\textrm{for}\quad 0\leq v:= \frac{1}{2} \int h^2 k(h)dh \ee
\noindent with $v$ characterizing the width (variance) of the kernel. For $v\neq 0$, we would approach a bit different coefficients than previously, which as discussed were optimal for the $\rho=\sum_i a_i f_i$ density.

Adding to a function its second derivative times a positive value corresponds to smoothing this function, analogously to evolution of diffusion equation ($\partial_t f=\partial_{xx}f$). Hence, the $v\to 0$ limit (to Dirac delta) intuitively corresponds to the sharpest possible fit.

\subsection{Normalization of density}
At least in theory, probabilistic density functions should integrate to 1. For polynomials and Fourier series it will be enforced by construction: all but the zero order $f_i$ will integrate to 0, hence, normalization of density can be obtained by just fixing the zero order coefficient. There are also situations, especially in machine learning, where the necessity of density integrating exactly to 1 is not crucial.

Let us now briefly focus on situations where density normalization is required, but cannot be directly enforced by construction. A trivial solution is just performing additional final normalization step: divide obtained $\rho_\textbf{a}$ by $\int\rho_\textbf{a}(\textbf{x})\,d\textbf{x}$.

However, more accurate values should be obtained by using the density normalization condition as constraint while the (\ref{min}) minimization, what can be done with the Lagrange multiplier method. Denoting
\be F_i:=\int f_i(\textbf{x}) d\textbf{x} \ee
\noindent the density normalization condition becomes $\sum_i a_i F_i = C$, where usually $C=1$, but generally can be also for example $C=\frac{1}{n}\sum_{\textbf{x}\in S} W(\textbf{x})$. For orthonormal set of functions we analogously get the following equations:

$$\sum_j a_j F_j = C,\qquad\ a_i = [f_i] + \lambda F_i\quad\textrm{for all}\ i.$$
\noindent Substituting to the first equation and transforming:
$$\sum_j ([f_j] + \lambda F_j) F_j = C $$
$$ \lambda = \left(C-\sum_j [f_j]F_j\right) / \sum_j (F_j)^2 $$
\be a_i = [f_i] + \frac{C-\sum_j [f_j]F_j}{\sum_j (F_j)^2}\, F_i \label{nor} \ee

The experiments from the right column of fig. \ref{overv} were made using this normalized formula for $C=1$.

\section{Polynomials, Fourier and Gaussian}
\noindent There will be now briefly discussed three natural examples of fitting with a linear combination, assuming weight $w=1$. As discussed, choosing this family as orthogonal allows to estimate the parameters independently. The first considered example are (Legendre) polynomials, the second is Fourier series, the last one are (Hermite) polynomials multiplied by $e^{-x^2/2}$.

\subsection{Fitting polynomial in a finite region}
A natural first application is parameterizing density function as a polynomial. The zeroth order function should be constant, but also needs to integrate to a finite value. Hence, we have to restrict to a finite region here (of volume $v<\infty$), preferable a range like $[-1,1]$ in fig. \ref{overv}, or a product of ranges (hyperrectangle) in a higher dimension. We will only use points inside this region to approximate (estimate) density inside this region with a polynomial - this region should be chosen as a relatively small one containing the behaviour of interest. All integrals and sums here are inside this finite region, for example by setting $w=0$ outside.

This zero order function is $f_0=1/\sqrt{v}$ to get normalization $\|f_0\|=1$. Its estimated coefficient is average of this function over the sample, which is always $a_0 = 1/\sqrt{v}$. Hence, the zero order estimation is just $\rho\approx a_0 f_0=1/v$ constant density (integrates to 1).

The following functions (polynomials) should be orthogonal to $f_0$, what means integration to 0 $(\int f_i\, d\textbf{x}=0)$, hence

\be \rho_\textbf{a} = \frac{1}{v} +\sum_{i\geq 1} a_i f_i \ee

\noindent always integrates to 1, probabilistic normalization is enforced for any choice of $a_i$ parameters ($i\geq 1$). However, we need to remember that such density sometimes may obtain negative values in the assumed region, like some red lines going below 0 in fig. \ref{overv}.

Orthonormal polynomials for $[-1,1]$ range and $w=1$ weight are known as Legendre polynomials. The first four are:

$$\frac{1}{\sqrt{2}},\ \sqrt{\frac{3}{2}}x,\ \sqrt{\frac{5}{8}}(3x^2-1),\ \sqrt{\frac{7}{8}}(5x^3-3x) $$

Thanks to orthogonality, we can independently ask about their coefficients. Finally, density estimation in $[-1,1]$ range with second order polynomial becomes:

\be \rho\approx \frac{1}{2}+\frac{3}{2}[x]x+\frac{5}{8}\left(3[x^2]-1\right)\left(3x^2-1\right) \label{square} \ee

\noindent where we have used linearity $[\alpha f+\beta g]=\alpha [f]+\beta [g]$, which makes it sufficient to calculate only averages of monomials over the sample - estimators of the moments. Such third order formula was used (and directly written with grouped $[x^i]$) in experiments presented in the left column of fig. \ref{overv}.

For a different interval, the Legendre polynomials should be properly shifted and rescaled. For $D$ dimensional case, if the region is a product of ranges (hyperrectangle), the orthogonal polynomials can be chosen as products of $D$ (rescaled) Legengre polynomials. Otherwise, there can be used Gram-Schmidt orthonormalization procedure to get an orthonormal base.

For example for $[-1,1]^D$ hypercube, the polynomial for density can be naturally chosen as: 
$$\rho(\textbf{x})=\frac{1}{2^D} + \sum_{i_1,\ldots,i_D} a_{i_1\ldots i_D}\,f_{i_1}(x_1)\cdot\ldots\cdot f_{i_D}(x_D) $$

The number of coefficient grow exponentially with dimension: fitting order $m$ polynomial in $D$ dimensions requires $m^D$ coefficients. However, some of them might be small and so neglected in the linear combination.

Before applying such fitting, it is crucial to properly choose the finite region of focus (like to the boundary between two classes), for example by normalizing it to $[-1,1]^D$ hypercube. This approach often brings some artifacts far from the sample, especially near the boundaries of the region. To reduce this effect, better effect can be obtained by using functions vanishing at these boundaries, like $\sin(j \pi x)$.

\subsection{Fourier series}
Like in Fig. \ref{2spir}, for some type of data Fourier base might be more appropriate, still requiring working on a finite region (bounded or torus), preferably a hyperrectangle. Its zero order term is again constant $1/v$ and guards the normalization. The orthonormal base for $[-1,1]$ range is formed by $\sin(j\pi x)$ and $\cos(j \pi x)$ functions, which do not influence the normalization:

\be \rho = \frac{1}{2}+\sum_{j\geq 1} a_j\sin(j\pi x) + b_j \cos(j \pi x)\ee
where $a_j=[\sin(j\pi x)]$, $b_j=[\cos(j\pi x)]$ are just the averages of this function over the sample. For $i$ up to $m$ in $D$ dimensions we need $(2m)^D$ coefficients here.

Observe that sine terms vanish at the boundaries of interval (hypercube) - using only them we can reduce artifacts at the boundaries.

If we need to work on a two-dimensional sphere, the complete orthonormal base of spherical harmonics can be used. In a more general case, the orthonormal base can be obtained by Gram-Schmidt orthonormalization procedure.

\subsection{Global fitting combination of vanishing functions}
Estimating density function with polynomials or Fourier series requires restriction to some finite region, what is a strong limitation and often brings some artifacts. To overcome it, we can for example use a combination of vanishing functions instead: with values dropping to zero while going away from the central point, for example in $e^{-x^2}$ or $e^{-|x|}$ or $1/|x|^K$ way. For the convenience of applying the (\ref{rho}) formula, it is preferred that this set of functions is orthogonal.

A well known example of such orthogonal family of functions are Hermite polynomial multiplied by $e^{-x^2/2}$. Denoting by $h_i$ as $i$-th Hermite polynomial, the following set of functions is orthonormal for $\langle f,g\rangle = \int_{\mathbb{R}} f(x)g(x) dx$ scalar product (weight $w=1$):

\be f_i(x) = \frac{1}{\sqrt{2^i\, i!\,\sqrt{\pi} }}\ h_i(x)\ e^{-x^2/2} \label{herm} \ee

This time it starts with $i=0$, Hermite polynomials $h_i$ for $i=0,1,2,3,4,5$ are correspondingly:

$$ 1,\ x,\ x^2-1,\ x^3-3x,\ x^4-6x^2+3,\ x^5-10x^3+15x $$

As the space has infinite volume here, we cannot use constant function in the considered base, which enforced $F_i = \int f_i dx =0$ for all but the constant orthogonal functions, making $\int \rho dx=1$ enforced by the zero order term, not changed by other parameters.

The odd order terms are asymmetric here, hence $F_i=0$ for them. However, even order terms integrate to nonzero values, hence the normalized formula (\ref{nor}) should lead to a bit better accuracy. It was used to obtain the right column of fig. \ref{overv}.\\

The used $e^{-x^2/2}$ is characteristic for Gaussian distribution with standard deviation equal 1. The remaining terms from the linear combination can slightly modify this standard deviation, however, the natural first approach for applying the discussed fitting is to perform normalization first. Like in ICA: first shift to the mean value to centralize the sample, then perform PCA (principal component analysis): find eignenvectors and eigenvalues of the covariance matrix, and rescale these directions correspondingly - getting normalized sample: with mean in 0 and unitary covariance matrix.

For such normalized sample we can start looking for distortions from the Gaussian distribution (often corresponding to a noise). Such sample is prepared to directly start fitting the $f_i$ from formula (\ref{herm}). However, it might be worth to first rotate it to emphasize directions with the largest distortion from the Gaussian distribution - which are suspected to carry the real signal. In ICA these directions are usually chosen as the ones maximizing kurtosis. Here we could experiment with searching for directions maximizing some $a_i$ coefficient instead: average of $f_i$ in the given direction over the sample. 

Observe that in contrast to ICA, the discussed approach also allows to reconstruct the modelled distortion of Gaussian distribution. It can also model distortion of different probability distributions.

Generally, while high dimensional cases might require relatively huge base, most of the coefficients might turn out practically zero and so can be removed from the sum for density. Like in ICA, it might be useful to search for the really essential distortions (both direction and order), which may represent the real signal. For example orthogonal base can be built by searching for a function orthonormal to the previously found, which maximizes own coefficient - then adding it to the base and so on.

\section{Some further possibilities}
\noindent The discussed approach is very general, here are some possibilities of extensions, optimizations for specific problems.

One line of possibilities is using a linear combination of different family of functions, for example obtained by Gram-Schmidt orthonormalization. For instance, while there was discussed perturbation of Gaussian distribution with Hermite polynomials, in some situations heavy tails might be worth to consider instead, vanishing for example like $e^{-|x|}$ or $1/|x|^K$.

While we have focused on global fitting in the considered region, there could be also considered local ones, for example as a linear combination of lattice of vanishing functions like wavelets. Or divide the space into (hyper)rectangles, fit polynomial in each of them and smoothen on the boundaries using a weighted average.

A different approach to local fitting is through modifying the weight $w$: focus on some point by using a weight vanishing while going away from this point. Then for example fit polynomial describing behavior of density around this point, getting estimation of derivatives of density function in this point, which then could be combined in a lattice by using some splines.\\

Another line of possibility is that while there were only discussed linear combinations as the family of density functions, it might be worth to consider also some different families, for example enforcing being positive and vanishing, like:
$$f_{\textbf{a}}(x) = e^{-\sum_i a_i\, x^i}\quad\textrm{or}\quad f_{\textbf{a}}(x) = \frac{1}{\sum_i a_i\, x^i}$$
where in both cases the polynomial needs to have even order and positive dominant coefficient. In the latter case it additionally needs to be strictly positive. Their advantage is that they are both vanishing and nonnegative, as a density function should be.\\

Finally, a wide space of possibilities is using the discussed method for different tasks than estimation of probability density integrating to 1. For example to predict probability of the current symbol basing on the previous occurrences, what is a crucial task in data compression. Or the sampled points may contain some weights, basing on which we might want to parameterize the density profile, for example of a chemical molecule~\cite{mole}. 

Very promising is also application for the clustering problem, especially the supervised situation: where we need to model a complex boundary between two labeled classes - using polynomials generalizes linear separators used in standard approaches, and here we can directly calculate the coefficients as just averages, eventually improving them later e.g. by backpropagation. Such classification can be done by assigning weights of opposite signs (or complex) to points from different classes, and look for the sign of the fitted density function (or argument). Its zero value manifold models the boundary. Often some classes are under-represented, what should be compensated by increasing weights of its representatives.

\section{Conclusions}
\noindent There was presented and discussed a very general and powerful, but still looking basic approach. It might be already known, but seems to be missed in the standard literature. Its discussed basic cases can be literally fitting a polynomial or Fourier series to the sample (in a finite region), or distortion from a standard probability distribution like Gaussian. There is a wide range of potential applications, starting with modelling of local density for data of various origin, testing PRNG, finding the real signal in a noise in analogy to ICA, reconstructing density of this distortion, or for various clustering problems. This article only introduces to this topic, discussing very basic possibilities, only mentioning some of many ways to expand this general approach of mean-square fitting to Dirac delta limit of smoothing the sample with a kernel.

\bibliographystyle{IEEEtran}
\bibliography{cites}
\end{document}